\documentclass[runningheads]{llncs}
\usepackage{booktabs}  
\usepackage{array}
\usepackage{bm}
\usepackage{multirow}
\usepackage{amssymb}
\usepackage{amsmath}
\usepackage[pagebackref=false,breaklinks=true,letterpaper=true,colorlinks,bookmarks=false]{hyperref}
\usepackage{cite}
\usepackage{url}
\usepackage{mathrsfs}
\usepackage{bbding}

\usepackage{graphicx}
\begin{document}
\title{CoTr: Efficiently Bridging CNN and Transformer for 3D Medical Image Segmentation}

%
\titlerunning{CoTr: Bridging CNN and Transformer for 3D Medical Image Segmentation}




\author{Yutong Xie$^{1,2}$\thanks{Y. Xie and J. Zhang contributed equally to this work.}, 
Jianpeng Zhang$^{1,2\star}$, 
Chunhua Shen$^{2}$,\\
Yong Xia\Envelope$^{1}$\\
}

\institute{$^1$ Northwestern Polytechnical University, China	
\\$^2$ The University of Adelaide, Australia
}

\maketitle              
\begin{abstract}
Convolutional neural networks (CNNs) have been the de facto standard for nowadays 3D medical image segmentation. The convolutional operations used in these networks, however, inevitably have limitations in modeling the long-range dependency due to their inductive bias of locality and weight sharing. Although Transformer was born to address this issue, it suffers from extreme computational and spatial complexities in processing high-resolution 3D feature maps.
In this paper, we propose a novel framework that efficiently bridges a {\bf Co}nvolutional neural network and a {\bf Tr}ansformer {\bf (CoTr)} for accurate 3D medical image segmentation.
Under this framework, the CNN is constructed to extract feature representations and an efficient deformable Transformer (DeTrans) is built to model the long-range dependency on the extracted feature maps.
Different from the vanilla Transformer which treats all image positions equally, our DeTrans pays attention only to a small set of key positions by introducing the deformable self-attention mechanism. Thus, the computational and spatial complexities of DeTrans have been greatly reduced, making it possible to process the multi-scale and high-resolution feature maps, which are usually of paramount importance for image segmentation.
We conduct an extensive evaluation on the Multi-Atlas Labeling Beyond the Cranial Vault (BCV) dataset that covers 11 major human organs. 
The results indicate that our CoTr leads to a substantial performance improvement over other CNN-based, transformer-based, and hybrid methods on the 3D multi-organ segmentation task.
Code is available at
  \def\UrlFont{\rm\small\ttfamily}
\url{https://github.com/YtongXie/CoTr} 

\keywords{3D Medical image segmentation \and Deformable self-attention \and CNN \and Transformer.}
\end{abstract}
\section{Introduction}

\begin{figure}[t]
	\begin{center}
		{\includegraphics[width=0.9\linewidth]{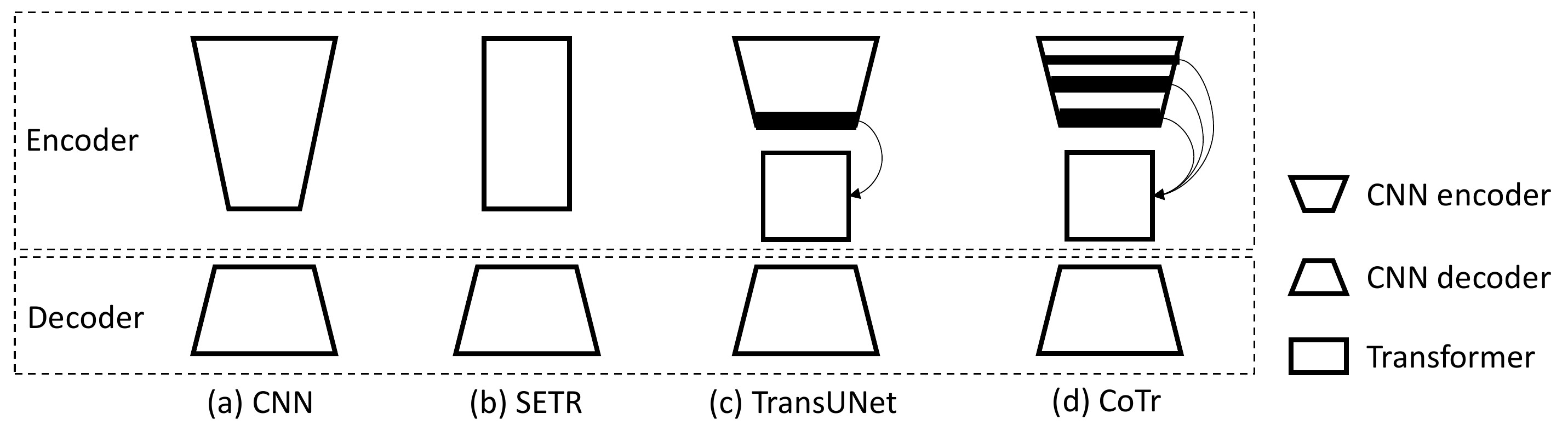}}
	\end{center}
	\vspace{-0.5cm}
	\caption{Comparison of different segmentation architectures. All of them have an encoder-decoder structure, but with different encoders. The encoder in CNN (a) is composed of multiple stacked convolutional layers.  The encoder in SETR (b) is purely formed from self-attention layers, $i.e.$, Transformer. The encoder in both TransUNet (c) and our proposed CoTr (d) are the hybrid of CNN and Transformer. Differently, TransUNet only processes the low-resolution feature maps from the last stage due to the high computation and spatial complexities. Thanks to the efficient design of Transformer, CoTr is able to process the multi-scale and high-resolution feature maps. 
	}	
	\label{fig:fig1}
\end{figure}

Image segmentation is a longstanding challenge in medical image analysis. 
Since the introduction of U-Net~\cite{unet}, fully convolutional neural networks (CNNs) have become the predominant approach to addressing this task~\cite{zhang2020block,Unet++,nnUnet,Yan, zhang2019light,zhang2020inter}. 
Despite their prevalence, CNNs still suffer from the limited receptive field and fail to capture the long-range dependency, due to the inductive bias of locality and weight sharing~\cite{inductive_bias}.
Many efforts have been devoted to enlarge a CNN's receptive field thus improve its ability to context modeling.
Yu \textit{et al.}~\cite{Fisher2016dilated} proposed the atrous convolution with an adjustable dilated rate, which shows superior performance in semantic segmentation~\cite{deeplabv3plus}. 
More straightforwardly, Peng~\textit{et al.}~\cite{peng2017largeKernel} designed large kernels to capture rich global context information.
Zhao~\textit{et al.}~\cite{PSPnet} employed the pyramid pooling at multiple feature scales to aggregate multi-scale global information.
Wang~\textit{et al.}~\cite{non_local} presented the non-local operations which is usually embedded at the end of encoder to capture the long-range dependency.
Although improving the context modeling to some extent, these models still have an inevitably limited receptive field, stranded by the CNN architecture.

Transformer, a sequence-to-sequence prediction framework, has a proven track record in machine translation and nature language processing~\cite{attention_is_all_you_need,Bert}, due to its strong ability to long-range modeling. The self-attention mechanism in Transformer can dynamically adjust the receptive field according to the input content, and hence is superior to convolutional operations in modeling the long-range dependency.

Recently, Transformer has been considered as an alternative architecture, and has achieved competitive performance on many computer vision tasks, like image recognition~\cite{ViT,DeiT}, semantic/instance segmentation~\cite{SETR,VisTR}, object detection~\cite{DETR,deformableDETR}, low-level vision~\cite{parmar2018image,chen2020pre}, and image generation~\cite{TransGAN}.
A typical example is the vision Transformer (ViT)~\cite{ViT}, which outperforms a ResNet-based CNN on recognition tasks but at a cost of using 300M data for training.
Since a huge training dataset is not always available, recent studies attempt to combine a CNN and a Transformer into a hybrid model. 
Carion~\textit{et al.}~\cite{DETR} employed a CNN to extract image features and a Transformer to further process the extracted features.
Chen \textit{et al.}~\cite{TransUNet} designed TransUNet, in which a CNN and a Transformer are combined in a cascade manner to make a strong encoder for 2D medical image segmentation. Although the design of TransUNet is interesting and the performance is good, it is challenging to optimize this model due to the existence of self-attention~\cite{attention_is_all_you_need}. 
First, it requires extremely long training time to focus the attention, which was initially cast to each pixel uniformly, on salient locations, especially in a 3D scenario.
Second, due to its high computational complexity, a vanilla Transformer~\cite{attention_is_all_you_need} can hardly process multi-scale and high-resolution feature maps, which play a critical role in image segmentation.

In this paper, we propose a hybrid framework that efficiently bridges {\bf Co}-nvolutional neural network and {\bf Tr}ansformer {\bf (CoTr)} for 3D medical image segmentation. CoTr has an encoder-decoder structure. In the encoder, a concise CNN structure is adopted to extract feature maps and a Transformer is used to capture the long-range dependency (see Fig.~\ref{fig:fig1}). Inspired by~\cite{dai2017deformable,deformableDETR}, we introduce the deformable self-attention mechanism to the Transformer. This attention mechanism casts attentions only to a small set of key sampling points, and thus dramatically reduces the computational and spatial complexity of Transformer. As a result, it is possible for the Transformer to process the multi-scale feature maps produced by the CNN and keep abundant high resolution information for segmentation.
The main contributions of this paper are three-fold: 
(1) we are the first to explore Transformer for 3D medical image segmentation, particularly in a computationally and spatially efficient way; 
(2) we introduce the deformable self-attention mechanism to reduce the complexity of vanilla Transformer, and thus enable our CoTr to model the long-range dependency using multi-scale features;
(3) our CoTr outperforms the competing CNN-based, Transformer-based, and hybrid methods on the 3D multi-organ segmentation task.

\begin{figure}[!t]
	\begin{center}
		{\includegraphics[width=1.0\linewidth]{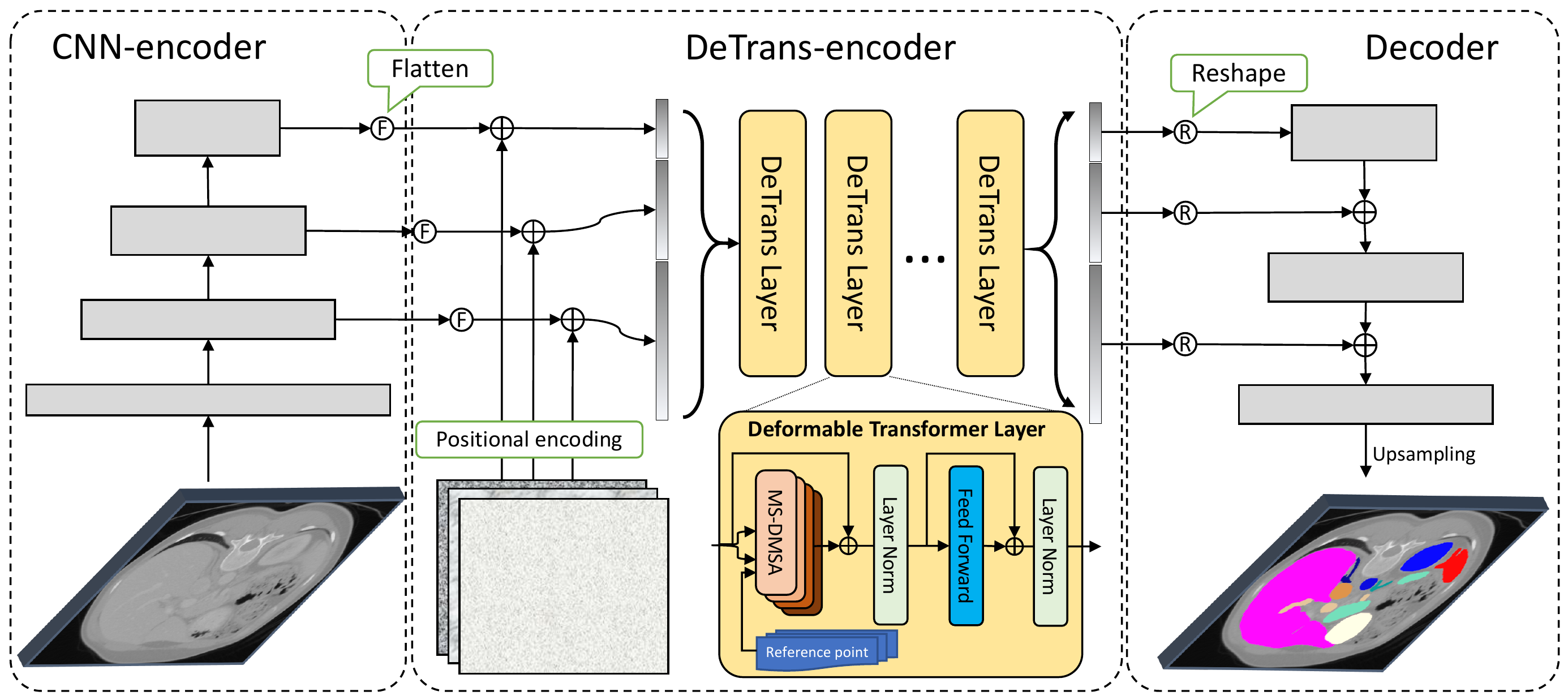}}
	\end{center}
	\vspace{-0.3cm}
	\caption{Diagram of CoTr: A CNN-encoder, a DeTrans-encoder, and a decoder. Gray rectangles: CNN blocks. Yellow rectangles: 3D deformable Transformer layers. 
	The CNN-encoder extracts multi-scale feature maps from an input image. 
	The DeTrans-encoder processes the flattened multi-scale feature maps that embedded with the positional encoding in a sequence-to-sequence manner. 
	The features with long-range dependency are generated by the DeTrans-encoder and fed to the decoder for segmentation. 
	}	
	\label{fig:framework}
\end{figure}

\vspace{-0.1cm}
\section{Materials}
\vspace{-0.1cm}
The Multi-Atlas Labeling \textbf{B}eyond the \textbf{C}ranial \textbf{V}ault (BCV) dataset~\footnote{\url{https://www.synapse.org/\#!Synapse:syn3193805/wiki/217789}} was used for this study. It contains 30 labeled CT scans for automated segmentation of 11 abdominal organs, including the spleen (Sp), kidney (Ki), gallbladder (Gb), esophagus (Es), liver (Li), stomach (St), aorta (Ao), inferior vena cava (IVC), portal vein and splenic vein (PSV), pancreas (Pa), and adrenal gland (AG).

\section{Methods}
CoTr aims to learn more effective representations for medical image segmentation via bridging CNN and Transformer. As shown in Fig.~\ref{fig:framework}, it consists of a CNN-encoder for feature extraction, a deformable Transformer-encoder (DeTrans-encoder) for long-range dependency modeling, and a decoder for segmentation.
We now delve into the details of each module.

\subsection{CNN-encoder}
The CNN-encoder $\mathcal{F}^{CNN}(\cdot )$ contains a Conv-IN-ReLU block and three stages of 3D residual blocks. The Conv-IN-ReLU block contains a 3D convolutional layer followed by an instance normalization (IN)~\cite{IN} and Rectified Linear Unit (ReLU) activation. The numbers of 3D residual blocks in three stages are three, three, and two, respectively. 

Given an input image $\bm{x}$ with a height of $H$, a width of $W$, and a depth ($i.e.$, number of slices) of $D$, the feature maps produced by $\mathcal{F}^{CNN}(\cdot )$ can be formally expressed as 
\begin{equation}
\{\bm{f}_{l}\}_{l=1}^{L}=\mathcal{F}^{CNN}_{l}(x;\bm{\Theta})\in \mathbb{R}^{C\times \frac{D}{{2}^{l}}\times \frac{H}{{2}^{l+1}}\times \frac{W}{{2}^{l+1}}}, 
\end{equation}
where $L$ indicates the number of feature levels, $\bm{\Theta}$ denotes the parameters of the CNN-encoder, and $C$ denotes the number of channels. 

\vspace{-0.1cm}
\subsection{DeTrans-encoder}
Due to the intrinsic locality of convolution operations, the CNN-encoder cannot capture the long-range dependency of pixels effectively. To this end, we propose the DeTrans-encoder that introduces the multi-scale deformable self-attention (MS-DMSA) mechanism for efficient long-range contextual modeling. The DeTrans-encoder is a composition of an input-to-sequence layer and $L_{D}$ stacked deformable Transformer (DeTrans) layers.

\vspace{+0.1cm}
\noindent{\textbf{Input-to-sequence Transformation.}} 
Considering that Transformer processes the information in a sequence-to-sequence manner, we first flatten the feature maps produced by the CNN-encoder $\{\bm{f}_{l}\}_{l=1}^{L}$ into a 1D sequence.
Unfortunately, the operation of flattening the features leads to losing the spatial information that is critical for image segmentation. To address this issue, we supplement the 3D positional encoding sequence $\{\bm{p}_{l}\}_{l=1}^{L}$ to the flattened $\{\bm{f}_{l}\}_{l=1}^{L}$.
For this study, we use sine and cosine functions with different frequencies~\cite{attention_is_all_you_need} to compute the positional coordinates of each dimension $pos$, shown as follows

\begin{equation}
\left\{\begin{matrix}
PE_{\#}(pos,2k)=sin(pos \cdot \upsilon) \\ 
PE_{\#}(pos,2k+1)=cos(pos \cdot \upsilon) 
\end{matrix}\right.
\end{equation}
where $\#\in\left \{D,H,W \right \}$ indicates each of three dimensions, $\upsilon=1/10000^{2k/\frac{C}{3}}$.
For each feature level $l$, we concatenate $PE_{D}$, $PE_{H}$, and $PE_{W}$ as the 3D positional encoding $\bm{p}_{l}$ and combine it with the flattened $\bm{f}_{l}$ via element-wise summation to form the input sequence of DeTrans-encoder.

\vspace{+0.1cm}
\noindent{\textbf{MS-DMSA Layer.}} 
In the architecture of Transformer, the self-attention layer would look over all possible locations in the feature map. It has the drawback of slow convergence and high computational complexity, and hence can hardly process multi-scale features.
To remedy this, we design the MS-DMSA layer that focuses only on a small set of key sampling locations around a reference location, instead of all locations.

Let $\bm{z}_q\in \mathbb{R}^{C}$ be the feature representation of query $q$ and $\hat{\bm{p}}_{q}\in [0,1]^{3}$
be the normalized 3D coordinate of the reference point. Given the multi-scale feature maps $\{\bm{f}_{l}\}_{l=1}^{L}$ that are extracted in the last $L$ stages of CNN-encoder, the feature representation of the $i$-th attention head can be calculated as

\begin{equation}
\begin{aligned}
{\rm head}_i = \sum_{l}^{L} \sum_{k}^{K} \Lambda (\bm{z}_q)_{ilqk} \cdot \Psi(\bm{f}_l)(\sigma_l(\hat{\bm{p}}_{q}) + \Delta _{\bm{p}_{ilqk}})
\end{aligned}
\end{equation}
where $K$ is the number of sampled key points, $\Lambda (\bm{z}_q)_{ilqk} \in [0,1]$ is the attention weight, $\Delta _{\bm{p}_{ilqk}}\in \mathbb{R}^{3}$ is the sampling offset of the $k$-th sampling point in the $l$-th feature level, and $\sigma_l(\cdot)$ re-scales $\hat{\bm{p}}_{q}$ to the $l$-th level feature. Following ~\cite{deformableDETR}, both $\Lambda (\bm{z}_q)_{ilqk}$ and $\Delta _{\bm{p}_{ilqk}}$ are obtained via linear projection over the query feature $\bm{z}_q$. Then, the MS-DMSA layer can be formulated as 

\begin{equation}
\begin{aligned}
&{\rm MS-DMSA}(\bm{z}_q, \{\bm{f}_l\}_{l=1}^{L}) = \Phi ({\rm Concat}({\rm head}_1, {\rm head}_2, ..., {\rm head}_H))
\end{aligned}
\end{equation}
where $H$ is the number of attention heads, and $\Phi(\cdot)$ is a linear projection layer that weights and aggregates the feature representation of all attention heads.

\vspace{+0.1cm}
\noindent{\textbf{DeTrans Layer.}} 
The DeTrans layer is composed of a MS-DMSA layer and a feed forward network, each being followed by the layer normalization~\cite{ba2016layer} (see Fig.~\ref{fig:framework}). The skip connection strategy~\cite{he2016deep} is employed in each sub-layer to avoid gradient vanishing. 
The DeTrans-encoder is constructed by repeatedly stacking DeTrans layers.

\subsection{Decoder}
The output sequence of DeTrans-encoder is reshaped into feature maps according to the size at each scale. The decoder, a pure CNN architecture, progressively upsamples the feature maps to the input resolution ($i.e.$, $D\times H \times W$) using the transpose convolution, and then refines the upsampled feature maps using a 3D residual block. Besides, the skip connections between encoder and decoder are also added to keep more low-level details for better segmentation. We also use the deep supervision strategy by adding auxiliary losses to the decoder outputs with different scales. The loss function of our model is the sum of the Dice loss and cross-entropy loss \cite{nnUnet,zhang2019light, Unet++}. 
More details on the network architecture gare in Appendix.

\subsection{Implementation details}
Following~\cite{nnUnet}, we first truncated the HU values of each scan using the range of $[-958, 327]$ to filter irrelevant regions, and then normalized truncated voxel values by subtracting 82.92 and dividing by 136.97. 
We randomly split the BCV dataset into two parts: 21 scans for training and 9 scans for test, and randomly selected 6 training scans to form a validation set, which just was used to select the hyper-parameters of CoTr. The final results on the test set are obtained by the model trained on all training scans.


In the training stage, we randomly cropped sub-volumes of size $48\times192\times192$ from CT scans as the input. To alleviate the over-fitting of limited training data, we employed the online data argumentation~\cite{nnUnet}, including the random rotation, scaling, flipping, adding white Gaussian noise, Gaussian blurring, adjusting rightness and contrast, simulation of low resolution, and Gamma transformation, to diversify the training set. 
Due to the benefits of instance normalization~\cite{IN}, we adopted the micro-batch training strategy with a small batch size of 2. 
To weigh the balance between training time cost and performance reward, CoTr was trained for 1000 epochs and each epoch contains 250 iterations.
We adopted the stochastic gradient descent algorithm with a momentum of 0.99 and an initial learning rate of 0.01 as the optimizer.
We set the hidden size in MS-DMSA and feed forward network to 384 and 1536, respectively, and empirically set the hyper-parameters $L_{D}=6$, $H=6$, and $K=4$.
Besides, we formed two variants of CoTr with small CNN-encoders, denoted as CoTr$^\ast$ and CoTr$^\dagger$. In CoTr$^\ast$, there is only one 3D residual block in each stage of CNN-encoder. In CoTr$^\dagger$, the number of 3D residual blocks in each stage of CNN-encoder is two.

In the test stage, we employed the sliding window strategy, where the window size equals to the training patch size. Besides, Gaussian importance weighting~\cite{nnUnet} and test time augmentation by flipping along all axes were also utilized to improve the robustness of segmentation. 
To quantitatively evaluate the segmentation results, we calculated the Dice coefficient scores (Dice) metric that measures the overlapping between a prediction and its ground truth.

\section{Results}

\begin{table}[!t]
\scriptsize
\renewcommand\arraystretch{1.25}
\caption{Dice scores of our CoTr and several competing methods on the BCV test set.  \textbf{CoTr}$^\ast$ and \textbf{CoTr}$^\dagger$ are two variants of CoTr with small CNN-encoders}
\label{tab:tab2}
\vspace{-0.5cm}
\begin{center}
\begin{tabular}{cccccccccccccm{0.7cm}<{\centering}}
\toprule[1.25pt]
\multirow{2}{*}{Methods} & \multirow{2}{*}{\begin{tabular}[c]{@{}c@{}}Param\\ (M)\end{tabular}} & \multicolumn{11}{c}{Organs}                                                & \multirow{2}{*}{Ave} \\ \cline{3-13} 
                         &                                                                      & Sp   & Ki   & Gb   & Es   & Li   & St   & Ao   & IVC  & PSV  & Pa   & AG   &                      \\ \midrule[1pt]
SETR (ViT-B/16-rand)~\cite{SETR}      & 100.5                                                                & 95.2 & 92.3 & 55.6 & 71.3 & 96.2 & 80.2 & 89.7 & 83.9 & 68.9 & 68.7 & 60.5 & 78.4                 \\ \hline
SETR (ViT-B/16-pre)~\cite{SETR}       & 100.5                                                                & 94.8 & 91.7 & 55.2 & 70.9 & 96.2 & 76.9 & 89.3 & 82.4 & 69.6 & 70.7 & 58.7 & 77.8                 \\ \hline
CoTr w/o CNN-encoder     & 21.9                                                                 & 95.2 & 92.8 & 59.2 & 72.2 & 96.3 & 81.2 & 89.9 & 85.1 & 71.9 & 73.3 & 61.0 & 79.8                 \\ \midrule[1pt]
CoTr w/o DeTrans         & 32.6                                                                 & 96.0 & 92.6 & 63.8 & 77.9 & 97.0 & 83.6 & 90.8 & 87.8 & 76.7 & 81.2 & 72.6 & 83.6                 \\ \hline
APSS~\cite{deeplabv3plus}                    & 45.5                                                                 & 96.5 & 93.8 & 65.6 & 78.1 & 97.1 & 84.0 & 91.1 & 87.9 & 77.0 & 82.6 & 73.9 & 84.3                 \\ \hline
PP~\cite{PSPnet}                       & 33.9                                                                 & 96.1 & 93.1 & 64.3 & 77.4 & 97.0 & 85.3 & 90.8 & 87.4 & 77.2 & 81.9 & 72.8 & 83.9                 \\ \hline
Non-local~\cite{non_local}                 & 32.8                                                                 & 96.3 & 93.7 & 64.6 & 77.9 & 97.1 & 84.1 & 90.8 & 87.7 & 77.2 & 82.1 & 73.3 & 84.1                 \\ \midrule[1pt]
TransUnet~\cite{TransUNet}                & 43.5                                                                 & 95.9 & 93.7 & 63.1 & 77.8 & 97.0 & 86.2 & 91.0 & 87.8 & 77.8 & 81.6 & 73.9 & 84.2                 \\ \hline
\textbf{CoTr}$^\ast$                  & 27.9                                                                 & 96.4 & 94.0 & 66.2 & 76.4 & 97.0 & 84.2 & 90.3 & 87.6 & 76.3 & 80.8 & 72.9 & 83.8                 \\ \hline
\textbf{CoTr}$^\dagger$                  & 36.9                                                                 & 96.2 & 93.8 & 66.5 & 78.6 & 97.1 & 86.9 & 90.8 & 87.8 & 77.7 & 82.8 & 73.2 & 84.7                 \\ \hline
\textbf{CoTr}                     & 41.9                                                                 & 96.3 & 93.9 & 66.6 & 78.0 & 97.1 & 88.2 & 91.2 & 88.0 & 78.1 & 83.1 & 74.1 & \textbf{85.0}                 \\ \bottomrule[1.25pt]
\end{tabular}
\end{center}
\label{tab:table1}
\vspace{-0.7cm}
\end{table}

\noindent{\textbf{Comparing to models with only Transformer encoder.}} 
We first evaluated our CoTr against two variants of the state-of-the-art SEgmentation Transformer (SETR)~\cite{SETR}, which were formed by using randomly initialized and pre-trained ViT-B/16~\cite{ViT} as the encoder. We also compared to a variant of CoTr that removes the CNN-encoder (CoTr w/o CNN-encoder). To ensure an unprejudiced comparison, all models use the same decoder.
The segmentation performance of these models is shown in Table~\ref{tab:table1}, from which three conclusions can be drawn.
First, although the Transformer architecture is not limited by the type of input images, the ViT-B/16 pre-trained on 2D natural images does not work well on 3D medical images. 
The suboptimal performance may be attributed to the domain shift between 2D natural images and 3D medical images. 
Second, `CoTr w/o CNN-encoder' has about 22M parameters and outperforms the SETR with about 100M parameters. We believe that a lightweight Transformer may be more friendly for medical image segmentation tasks, where there is usually a small training dataset. 
Third, our CoTr$^\ast$ with comparable parameters significantly outperforms `CoTr w/o CNN-encoder', improving the average Dice over 11 organs by 4\%. It suggests that the hybrid CNN-Transformer encoder has distinct advantages over the pure Transformer encoder in medical image segmentation.

\vspace{+0.1cm}
\noindent{\textbf{Comparing to models with only CNN encoder.}} 
Then, we compared CoTr against a variant of CoTr that removes the DeTrans-encoder (CoTr w/o DeTrans) and three CNN-based context modeling methods, $i.e.$, the Atrous Spatial Pyramid Pooling (ASPP)~\cite{deeplabv3plus} module, pyramid parsing (PP)~\cite{PSPnet} module, and Non-local~\cite{non_local} module. 
For a fair comparison, we used the same CNN-encoder and decoder but replaced our DeTrans-encoder with ASPP, PP, and Non-local modules, respectively.
The results in Table~\ref{tab:table1} shows that our CoTr elevates consistently the segmentation performance over `CoTr w/o DeTrans' on all organs and improves the average Dice by 1.4\%. It corroborates that our CoTr using a hybrid CNN-Transformer encoder has a stronger ability than using a pure CNN encoder to learn effective representations for medical image segmentation.
Moreover, comparing to these context modeling methods, our Transformer architecture contributes to more accurate segmentation. 

\vspace{+0.1cm}
\noindent{\textbf{Comparing to models with hybrid CNN-Transformer encoder.}} 
We also compared CoTr to other hybrid CNN-Transformer architectures like TransUNet~\cite{TransUNet}. To process 3D images directly, we extended the original 2D TransUNet to a 3D version by using 3D CNN-encoder and decoder as done in CoTr. 
We also set the number of heads and layers of Transformer in 3D TransUNet to be the same as our CoTr. 
It shows in Table~\ref{tab:table1} that CoTr steadily beats TransUNet in the segmentation of all organs, particularly for the gallbladder and pancreas segmentation. 
Even with a smaller CNN-encoder, CoTr$^\dagger$ still achieves better performance than TransUnet in the segmentation of seven organs. 
The superior performance owes to the deformable mechanism in CoTr that makes it possible to process high-resolution and multi-scale feature maps due to the reduced computational and spatial complexities.

\vspace{+0.1cm}
\noindent{\textbf{Computational Complexity.}} 
The proposed CoTr was trained using a workstation with a NVIDIA GTX 2080Ti GPU and the Pytorch software packages. It took about 2 days for training, and less than 30ms to segment a volume of size $48\times192\times192$.

\section{Discussion on Hyper-parameter Settings}

\begin{figure}[!t]
	\begin{center}
		{\includegraphics[width=1.0\linewidth]{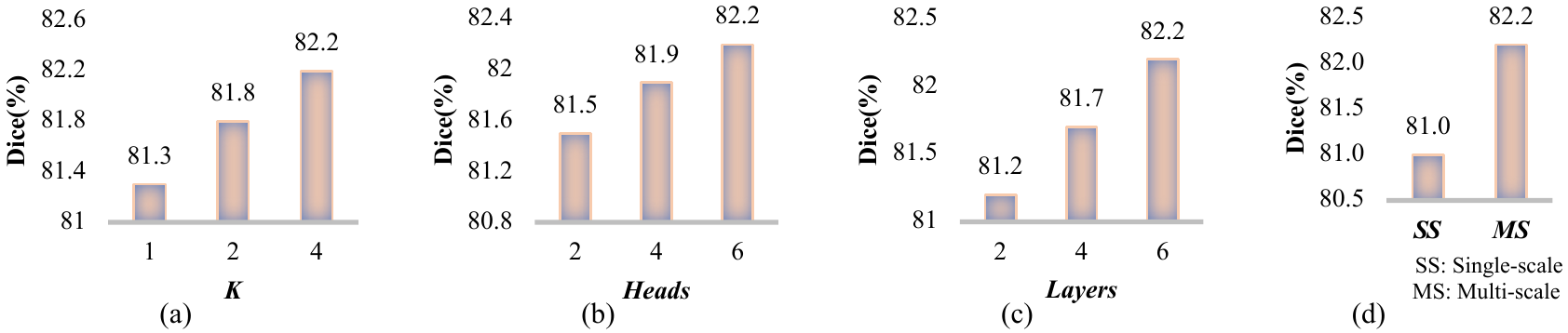}}
	\end{center}
	\vspace{-0.6cm}
	\caption{Average Dice over all organs obtained on the validation set versus (a) the number of sampled key points $K$, (b) number of heads $H$, and (c) number of DeTrans layers $L_{D}$, and (d) Average Dice obtained by our CoTr using, respectively, single-scale and multi-scale feature maps on the validation set.
	}	
	\label{fig:fig2}
	\vspace{-0.2cm}
\end{figure}

In the DeTrans-encoder, there are three hyper-parameters, $i.e.$, $K$, $H$, and $L_{D}$, which represent the number of sampled key points, heads, and stacked DeTrans layers, respectively. To investigate the impact of their settings on the segmentation, we set $K$ to 1, 2, and 4, set $H$ to 2, 4, and 6, and set $L_{D}$ to 2, 4, and 6. In Fig.~\ref{fig:fig2} (a-c), we plotted the average Dice over all organs obtained on the validation set versus the values of $K$, $H$, and $L_{D}$. 
It shows that increasing the number of $K$, $H$, or $L_{D}$ can improve the segmentation performance. 
To demonstrate the performance gain resulted from the multi-scale strategy, we also attempted to train CoTr with single-scale feature maps from the last stage. The results in Fig.~\ref{fig:fig2} (d) show that using multi-scale feature maps instead of single-scale feature maps can effectively improve the average Dice by 1.2\%.

\section{Conclusion}
In this paper, we propose a hybrid model of CNN Transformer, namely CoTr, for 3D medical image segmentation. In this model, we design the deformable Transformer (DeTrans) that employs the deformable self-attention mechanism to reduce the computational and spatial complexities of modelling the long-range dependency on multi-scale and high-resolution feature maps. Comparative experiments were conducted on the BCV dataset. The superior performance of our CoTr over both CNN-based and vanilla Transformer-based models suggests that, via combining the advantages of CNN and Transformer, the proposed CoTr achieves the balance in keeping the details of low-level features and modeling the long-range dependency. 
As a stronger baseline, our CoTr can be extended to deal with other structures (\textit{e.g.}, brain structure or tumor segmentation) in the future.


{
	\bibliographystyle{splncs04}
	\bibliography{egbib}
}

\section{Appendix}
\subsection{Detailed network architecture}
\begin{figure}[!h]
	\begin{center}
		{\includegraphics[width=1.0\linewidth]{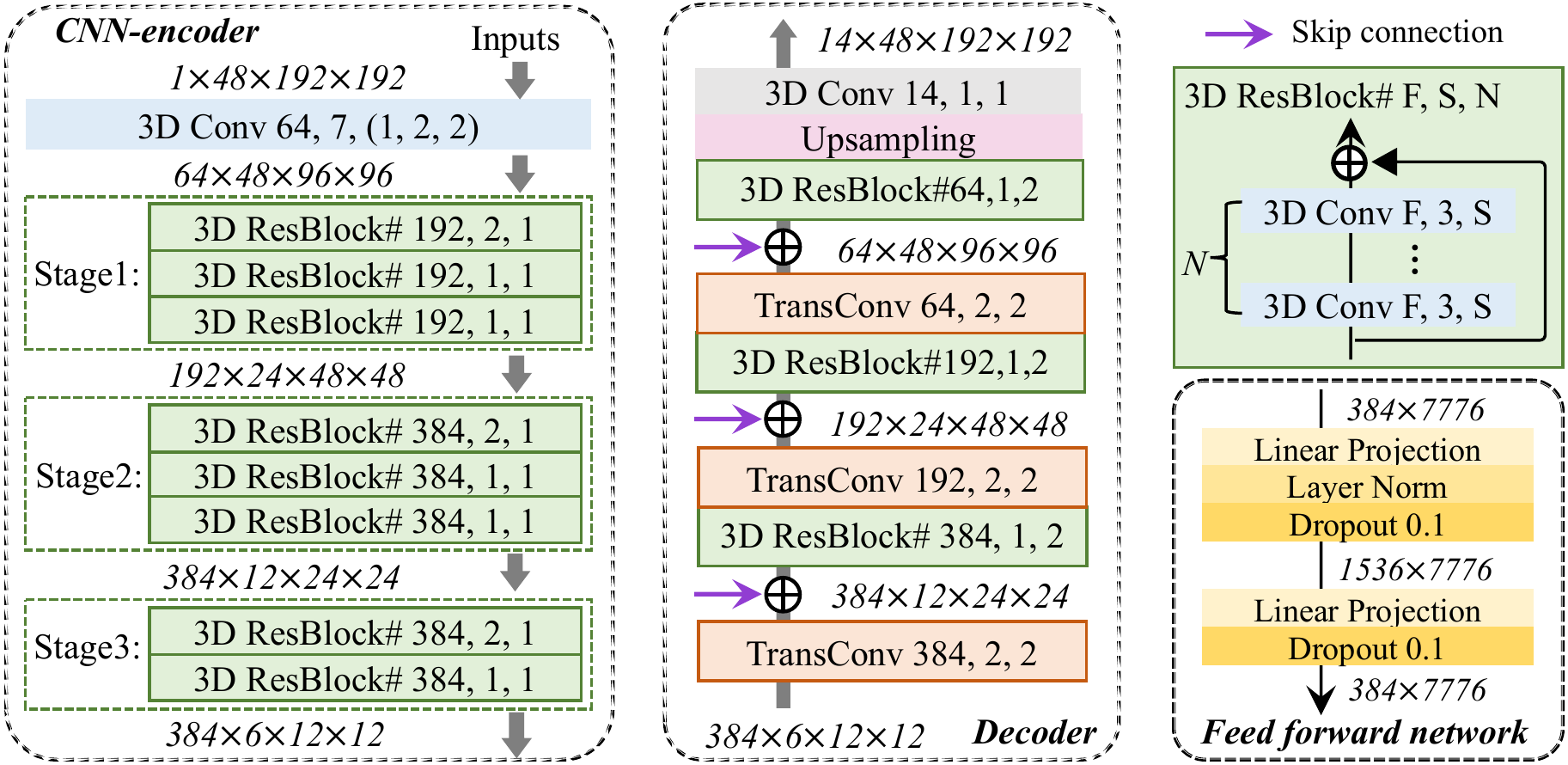}}
	\end{center}
	\vspace{-0.6cm}
	\caption{Detailed architecture of CNN-encoder, decoder and feed forward network. Blue ‘Conv’: Conv-In-Relu block that contains a 3D convolutional layer followed by an instance normalization (IN) layer~\cite{IN} and ReLU activation; Gray ‘Conv’: a 3D convolutional layer; Orange ‘TransConv’: 3D transposed convolutional layer. Note that the numbers in each Conv block / layer indicate the number of filters, kernel size, and stride, respectively. The numbers in each residual block indicate the number of filters, stride, and Conv blocks, respectively.
	}	
	\label{fig:Supp1}
\end{figure}

Fig.~\ref{fig:Supp1} shows the architecture of CNN-encoder, decoder and feed forward network in Detrans-encoder. 
It consists of a Conv-In-Relu and three stages of 3D residual blocks. The numbers of 3D residual blocks are three, three, and two in three stages, respectively.
The decoder contains four upsampling modules. Each of first three modules has a TransConv layer followed by a residual block, and a pixel-wise summation with the
corresponding feature maps from the encoder and the TransConv layer. The last module comprises of an Upsampling layer followed by a 1 × 1 Conv layer that
maps the 64-channel feature maps to the desired number of classes. 
The feed forward network in Detrans-encoder has two linear projection layers. The first layer is followed by a layer normalization layer and a Dropout layer. The second layer is followed by a Dropout layer.

\subsection{Loss function}
We jointly use the Dice loss and cross-entropy loss for optimization, which is popular in many medical image segmentation applications and has achieved prominent success~\cite{unet, Unet++, zhang2019light}. 
The loss function is formulated as 
\begin{equation}
\begin{aligned}
\small
\mathcal{L}=\frac{1}{c}\sum_{n=1}^{c}\left \{-\frac{2\sum \tilde{\bm{y}}^{n}\bm{y}^{n}+ \varepsilon}
{\sum(\tilde{\bm{y}}^{n}+\bm{y}^{n})+ \varepsilon} -\mathbb{E}\left [\bm{y}^{n}log\tilde{\bm{y}}^{n} \right ] \right \}
\end{aligned}
\end{equation}
where the first item is the soft Dice loss, the second item is the cross-entropy loss, the prediction and ground truth are denoted by $\tilde{y}$ and $y$, respectively, $\mathbb{E}$ is the expectation operation, $\varepsilon$ is a smoothing factor, and $c$ is the number of categories. To speed up convergence and alleviate the vanishing gradient problem, we also use the deep supervision strategy that adds auxiliary losses to the decoder outputs with different resolutions. The total loss function is the sum of the losses at all resolutions.

\subsection{Visualization}
\begin{figure}[!t]
	\begin{center}
		{\includegraphics[width=1.0\linewidth]{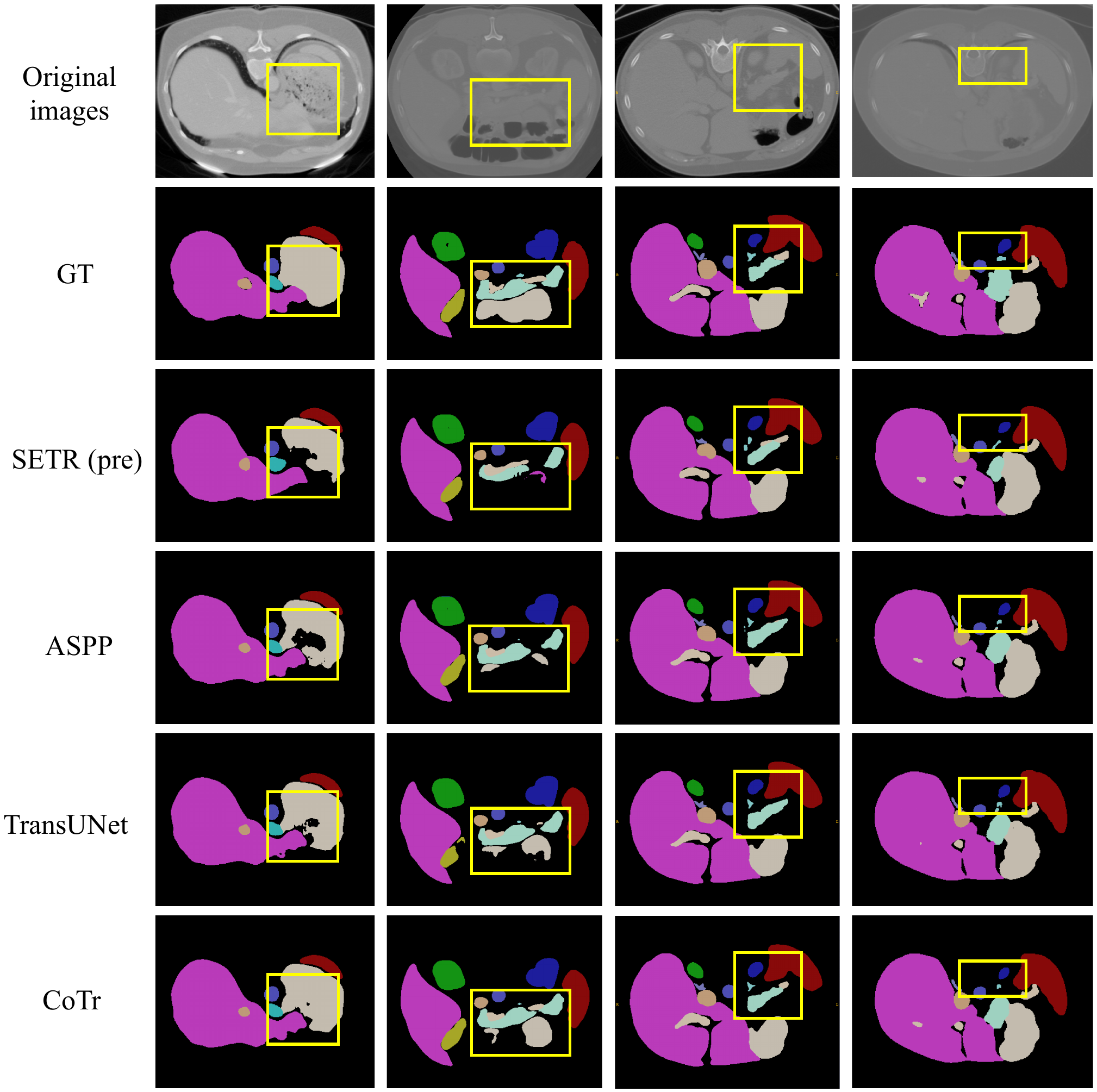}}
	\end{center}
	\vspace{-0.6cm}
	\caption{Visualization of segmentation results of four cases. The regions in yellow rectangles indicate our superiority. Each type of organs are denoted by a unique color.
	}	
	\label{fig:Supp2}
	\vspace{-0.5cm}
\end{figure}

The segmentation results produced by (1) SETR with pre-trained ViT-B/16, (2) replacing DeTrans-encoder with ASPP module, (3) 3D TransUNet, and (4) our CoTr, were visually compared in Fig.~\ref{fig:Supp2}. We can see that: 1) comparing to the pure Transformer encoder method (SETR) and pure CNN encoder method (ASPP), our CoTr with the hybrid CNN-Transformer encoder is able to produce the segmentation results that are more similar to the ground truth, and 2) our CoTr are more likely to produce less false positives compared to TransUNet, which confirms the superiority of our 3D deformable Transformer over vanilla Transformer.

\end{document}